\documentclass[12pt, a4paper]{article}
\usepackage[utf8]{inputenc}
\usepackage[left=2cm, right=2cm, bottom=2cm, top=2cm]{geometry}
\usepackage{xcolor}
\usepackage{svg}
\usepackage{booktabs}
\usepackage{graphicx}
\usepackage{caption}
\usepackage{subcaption}
\usepackage{amsmath, amsfonts, amssymb}
\usepackage{siunitx}

\DeclareSIUnit{\angstrom}{\textup{\AA}}
\usepackage{mathtools}
\usepackage{csvsimple}
\usepackage{authblk}
\usepackage{enumitem}
\usepackage[autostyle=true]{csquotes}
\usepackage{xparse}
\usepackage{bm}
\usepackage{pgfkeys, pgfsys, pgfcalendar}  

\definecolor{amethyst}{rgb}{0.6, 0.4, 0.8}
\usepackage[colorlinks, citecolor=blue, linkcolor=teal, urlcolor=amethyst]{hyperref}
\usepackage[capitalise, nameinlink]{cleveref}

\newcommand*{\suppnameref}[1]{\hyperref[{#1}]{Supplementary material, \nameref*{#1}}}

\NewDocumentCommand{\qnameref}{sm}{%
  \enquote{%
    \IfBooleanTF{#1}{\nameref*{#2}}{\nameref{#2}}%
  }%
}
\NewDocumentCommand{\sfref}{sm}{\IfBooleanTF{#1}{\nameref*{#2}}{\nameref{#2}}%
}
\NewDocumentCommand{\suppqnameref}{sm}{%
  \enquote{%
    \IfBooleanTF{#1}{\suppnameref*{#2}}{\suppnameref{#2}}%
  }%
}

\newlist{steps}{enumerate}{10}
\setlist[steps]{label=\arabic*., ref=\arabic*}

\makeatletter
\if@cref@capitalise
\crefname{stepsi}{Step}{Steps}
\else
\crefname{stepsi}{step}{steps}
\fi
\makeatother

\usepackage{biblatex}
\date{}

\begin{document}

\title{AAVDiff: Experimental Validation of Enhanced Viability and Diversity in Recombinant Adeno-Associated Virus (AAV) Capsids through Diffusion Generation}

\author[a]{\textsuperscript{1}Lijun Liu}
\author[b]{\textsuperscript{1}Jiali Yang}
\author[a]{\textsuperscript{1}Jianfei Song}
\author[b]{Xinglin Yang}
\author[b]{Lele Niu}
\author[b]{Zeqi Cai}
\author[a]{Hui Shi}
\author[c]{Tingjun Hou}
\author[a]{\textsuperscript{*}Chang-yu Hsieh}
\author[b]{\textsuperscript{*}Weiran Shen}
\author[b,d]{\textsuperscript{*}Yafeng Deng}

\affil[a]{ Hangzhou Carbonsilicon AI Technology Co., Ltd, Hangzhou 310018, Zhejiang, China
\newline
\texttt{\{liulijun, songjianfei, shihui, dengyafeng\}@carbonsilicon.ai, kimhsieh@zju.edu.cn}}

\affil[b]{ OBiO Technology ( Shanghai ) Corp.,Ltd, Shanghai 201318, China
\newline
\texttt{\{yjl, yxl, nll2429, czq2585, weiran.shen\}@obiosh.com}}

\affil[c]{ College of Pharmaceutical Science, Zhejiang University, Hangzhou 310058, Zhejiang, China
\newline
\texttt{tingjunhou@zju.edu.cn}}

\affil[d]{ Department of Automation, Tsinghua University, Beijing 100084, China}

\maketitle

\begin{abstract}
Recombinant adeno-associated virus (rAAV) vectors have revolutionized gene therapy, but their broad tropism and suboptimal transduction efficiency limit their clinical applications. To overcome these limitations, researchers have focused on designing and screening capsid libraries to identify improved vectors. However, the large sequence space and limited resources present challenges in identifying viable capsid variants. In this study, we propose an end-to-end diffusion model to generate capsid sequences with enhanced viability. Using publicly available AAV2 data, we generated 38,000 diverse AAV2 viral protein (VP) sequences, and evaluated 8,000 for viral selection. The results attested the superiority of our model compared to traditional methods. Additionally,  in the absence of AAV9 capsid data, apart from one wild-type sequence, we used the same model to directly generate a number of viable sequences with up to 9 mutations. we transferred the remaining 30,000 samples to the AAV9 domain. Furthermore, we conducted mutagenesis on AAV9 VP hypervariable regions VI and V, contributing to the continuous improvement of the AAV9 VP sequence. This research represents a significant advancement in the design and functional validation of rAAV vectors, offering innovative solutions to enhance specificity and transduction efficiency in gene therapy applications.
\end{abstract}

\section*{Introduction}
\label{sec:intro}

Recombinant Adeno-associated virus vectors (rAAV) have emerged as crucial components in the field of gene therapy. Since 2017, there have been six new gene therapy products approved, and over 2000 pipelines are registered, underscoring the significance of rAAV in clinical applications \cite{shen2022raav}. However, all six of the approved products employ capsid sequences that originate from wild-type viruses found in natural resources. Although these capsids derived from wild-type viruses exhibit broad tropism during treatment, rendering them non-specific in targeting pathogenic cells, their efficiency in transduction and gene expression within target cells is suboptimal. Consequently, they prove inadequate for the treatment of diseases primarily affecting specific tissues such as the central nervous system, muscle, and heart. Thus, there exists a consensus within the scientific community to advance the development of enhanced vectors with improved specificity and transduction efficiency.

Several methods have been established to design and evaluate novel capsids, and one promising approach is the design and screening of capsid libraries. This approach involves the creation of a pool of capsid-encoding DNA, which can be designed either rationally or randomly \cite{munch2013displaying}\cite{zhong2008next}\cite{muller2003random}\cite{lin2007recombinant}\cite{li2020engineering}. These DNA sequences are integrated into VP expression cassettes to facilitate vector production. The plasmid-to-cell ratio is meticulously fine-tuned during vector manufacturing to promote the production of a specific capsid variant, which encapsulates its own genome \cite{schmit2020cross},\cite{nonnenmacher2015high}. Subsequently, this pool of vectors is generated and injected into a selection model as a mixture. The resulting DNA signal delivered to the target cells is then retrieved and sequenced, representing the capsid variants that effectively transduce the target cells.

A common approach in library design involves either the insertion of a randomized peptide or the random mutation of specific amino acids within a tolerant domain \cite{muller2003random}. Notably, a prominent variant that has emerged through this library screening approach is PHP.B \cite{khasa2021analytical}. However, its ability to cross the blood-brain barrier (BBB) is not maintained during the translation of studies from mice to humans since the receptor for the PHP.B variant in brain microvascular endothelial cells is specific to particular mouse strains \cite{hordeaux2019gpi}. Several studies utilize a similar strategy by inserting 7 random amino acids within the capsid hypervariable VIII region; however, a clear frontrunner for clinical use has not yet emerged. Although the AAV VP coding region consists of approximately 720 amino acids, the insertion of 7 amino acids represents only a small fraction. Mutations in larger regions show promise in addressing diverse requirements. Nevertheless, even with 7 amino acid random mutations, the number of variants in the library can impose limitations on bacterial transformation, clone numbers, and vector manufacturing. Furthermore, the number of dosing iterations in a selection model becomes constrained. Therefore, it is crucial to explore broader mutational landscapes while ensuring the library sizes remain manageable.

Not all sequences resulting from capsid mutation can effectively express protein, assemble into a particle, and efficiently encapsulate their genome like the wild-type sequence. As the number of mutations in a VP increases, the sequence search space expands exponentially, making it impossible to filter through experimental means, resulting in a decreased likelihood of successful capsid packaging. The development of algorithms that establish a correlation between capsid DNA sequences and packaging efficiency is of utmost importance \cite{kelsic2019challenges}. Furthermore, low yield resulting from unfavorable physical and chemical properties can impede the clinical and commercialization potential. Attaining efficient and targeted transduction of specific cells poses a significant challenge in capsid engineering.

In order to overcome these challenges, researchers have utilized generative algorithms to design and predict the viability of viral vectors, specifically focusing on vector fitness. The most recent approach \cite{bryant2021deep} entails training a binary classifier using a substantial amount of capsid data to ascertain the viability of a given sequence. Subsequently, random sampling is conducted within a mutation subspace that has been randomly partitioned. Samples that are classified as viable by the binary classifier are retained, while non-viable samples are discarded. This iterative filtering process is used to select a collection of capsid sequences with potentially viable properties. The constructed capsid sequence collection using this method has a higher proportion of viable compared to the collection constructed by random mutation. Nevertheless, the ratio of viable sequences is heavily influenced by the performance of the trained binary classifier. Moreover, due to the vast number of possible combinations resulting from sequence mutations (excluding insertions), the combinatorial count reaches $2^{seq_len}$, where $seq_len$ represents the sequence length. This renders it impractical to complete the filtering process within a reasonable timeframe given the extensive range of choices. Consequently, during the implementation phase, it is imperative to randomly partition a subspace from this dataset and subsequently conduct the filtering process. Considering that the proportion of genuinely viable sequence samples in the overall sequence space is exceedingly low, there is a high probability of overlooking potential sequences when partitioning the subspace. Therefore, to address this issue, we combined the classification and filtering stages by introducing an end-to-end, diffused-based generative model that can effectively generate a higher proportion of viable sequences. Moreover, this model functions as a generative model that generates sequences by following the gradient direction to identify viable samples during the generation process. Consequently, this generation method enables the sampling of a greater number of potential viable samples within the designated timeframe.In this study, we employed the model trained using publicly available data on AAV2 to generate a collection of 38,000 highly diverse AAV2 VP sequences. Out of these, 8,000 sequences were randomly chosen and subjected to evaluation for their viral selection values through DNase-resistant capsid assembly testing, which revealed a significant improvement in performance compared to traditional methods \cite{bryant2021deep}. Moreover, the availability of viable data generated from mutations on wild-type capsids of various serotypes is severely limited, and the synthesis process is both time-consuming and expensive. Therefore, the direct generation of data with multiple mutation sites while preserving capsid viability during the mutation process on new serotypes would greatly expedite research in the field of capsids. Building upon this, we transferred the remaining 30,000 samples from the initial 38,000 AAV2-generated sequences to the corresponding domain of AAV9. These sequences will be synthesized into a vector library to assess their actual survival rate. Encouragingly, we observed positive results in terms of yield, and when the number of mutation sites reached 9, we achieved a relatively high proportion of viable samples. 

In conclusion, the advancement of rAAV vectors with improved specificity, transduction efficiency, and delivery mechanisms presents tremendous potential for gene therapy research. The design and screening of capsid libraries, complemented by generative algorithms, offer a dynamic approach to overcome the limitations of wild-type capsids, bringing us closer to the development of highly efficient and targeted gene therapy vectors. This study represents a significant progression in the field of viral vector design and functional validation, providing innovative solutions to the challenges encountered in gene therapy.

\section*{Experiments}
\paragraph{Experiment 1} In order to verify the ability of the diffusion model in AAV capsid sequence design, we performed the following experiments:
\begin{itemize}
     \item{1. Experiment on AAV2 HVR VIII}
     The diffusion model was trained using the data provided in the references\cite{bryant2021deep},\cite{ogden2019comprehensive}. After deduplicating the generated sequences and removing samples that overlapped with the training set, a collection of approximately 38,000 samples remained. Out of these, 8,000 samples were randomly selected for biological activity testing specifically targeting AAV2.
     \item{2. Experiment on Region VIII on AAV9}
     Proceeding with the remaining 30,000 samples from the previously generated sequences, activity experiments were conducted targeting AAV9. The specific methodology involved the direct replacement of the sequence fragment corresponding to region VIII of AAV9 with the generated sequence, followed by subsequent biological activity experiments.
\end{itemize}
\paragraph{Experiment 2}    
In order to explore the mutation fitness of multiple hypervariable regions on AAV9 serotypes, saturated single mutants were constructed on regions IV, V, and VIII.

\section*{Results}

\subsection*{Sequence generated by diffusion model}\label{subsec:intro_diffpalm}

To evaluate the reliability of the sequences generated by the diffusion model, we analyze them from two perspectives. The first perspective involves observing the relationship between the generated sequences and the training set. The second perspective entails experimentally validating the viability of the generated sequences.

\begin{figure}[h!]
\centering
\includegraphics[width=0.9 \textwidth]{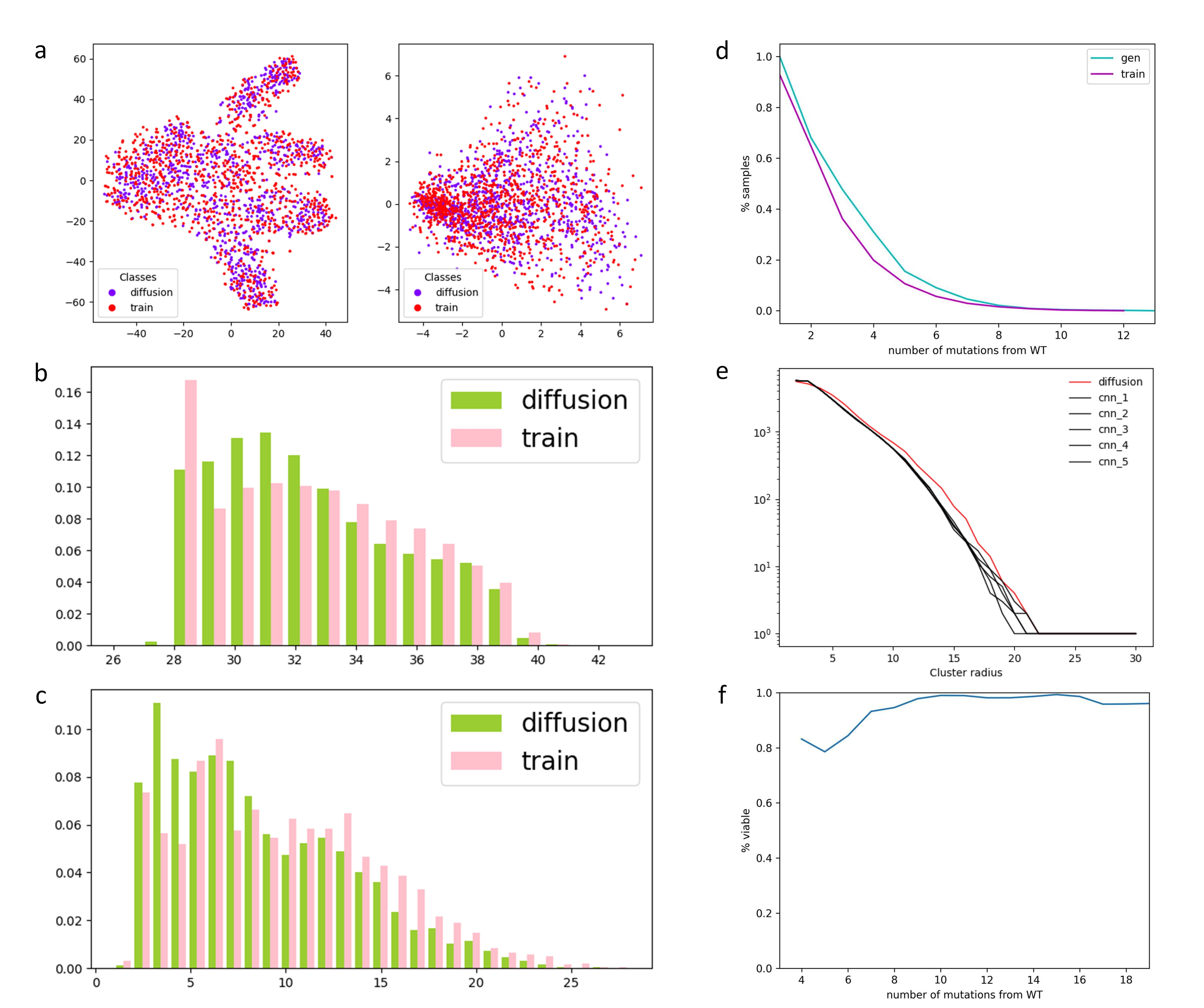}
\caption{\textbf{a:} Distribution of features of sequences generated by the diffusion model compared to the training set. The left graph represents the feature distribution after dimensionality reduction using t-SNE, while the right graph represents the feature distribution after dimensionality reduction using PCA. Class 1 represents the generated sequences, while classes 2 and 3 represent sequences from the training set. \textbf{b:} Distribution of sequence lengths for sequences generated by the diffusion model compared to the training set. The x-axis represents the length of the sequences, and the y-axis represents the frequency. The green color represents the generated sequences, while the other two colors represent sequences from the training set. \textbf{c:} Distribution of the number of mutation sites for sequences generated by the diffusion model compared to the training set. The x-axis represents the number of mutation sites, and the y-axis represents the frequency. The green color represents the generated sequences, while the other two colors represent sequences from the training set. \textbf{d:} Distribution of different lengths of consecutive insertions generated by the diffusion model. The x-axis represents the length of consecutive insertions, and the y-axis represents the proportion of samples. \textbf{e:} Distribution of the number of clusters for sequences generated by the diffusion model and the CNN model. The x-axis represents the clustering radius (sequences with a difference in mutation count within this radius are considered in the same cluster), and the y-axis represents the number of clusters. \textbf{f:} Proportion of viable samples for sequences generated by the diffusion model at different numbers of mutation sites. The x-axis represents different numbers of mutation sites, and the y-axis represents the proportion of viable samples.}
\label{fig:results_sgd}
\end{figure}

\paragraph{The relationship between the generated sequences and the training set}: The overlap between the feature space of the generated sequences and the training set can be observed from \cref{fig:results_sgd}a. \cref{fig:results_sgd}bdemonstrates the close match between the length distribution of the sequences generated by the generation model and the length distribution of the training set. Furthermore, the model generates sequences with shorter lengths, including those absent in the training set, such as at positions where the sequence length is 27. The distribution of the number of mutated positions generated by the model, as shown in \cref{fig:results_sgd}c, broadly covers all the numbers of mutated positions in the training set. In previous approaches to designing AAV capsid sequences, the design space was limited by the insertion or replacement of one amino acid between adjacent residues. However, the diffusion model does not impose such restrictions and allows for the insertion of one or more amino acids at specific positions. \cref{fig:results_sgd}d indicates that the sequences generated by the model exhibit a higher frequency of continuous insertions based on the WT sequence when compared to the training set. This can be attributed to our data augmentation approach, which incorporates continuous deletions and insertions between mutation positions. The method proposed by Dyno Therapeutics \cite{bryant2021deep} for designing highly active sequences involves selecting a seed sequence at a distance of k from the WT sequence. Subsequently, a single mutation is applied to the seed sequence to generate sequences that are at a distance of k+1 from the WT sequence. However, this greedy design approach restricts the diversity of the final sequences. Thus, \cref{fig:results_sgd}e illustrates the disparities in the number of clusters between the diffusion model and the CNN model \cite{bryant2021deep} at varying clustering radii. Greater sequence diversity is indicated by a higher number of clusters. The sequences designed by the diffusion model demonstrate slightly higher diversity when compared to those designed by the CNN model.

\paragraph{Performance of the generated sequences in terms of viability}:For the biological viability experiments on the AAV2 region VIII, a random selection of 8000 samples was chosen from the generated sequences. \cref{fig:results_sgd}f illustrates the proportion of viable samples among the sequences generated by the diffusion model, considering different numbers of mutations. The proportion of viable samples exceeds 90\% when the number of mutations ranges from 7 to 20, as evident from the observations. Additional detailed information can be found \cref{tab:dataset_AAV2}. The proportion of viable samples is approximately 80\% for mutation numbers ranging from 4 to 6. These results clearly demonstrate the robust capability of our model to generate viable sequences.

\subsection*{Performance of the diffusion model-generated sequences when transferred to the AAV9 serotype:}\label{subsec:intro_diffpalm}

\begin{figure}[h!]
\centering
\includegraphics[width=0.7 \textwidth]{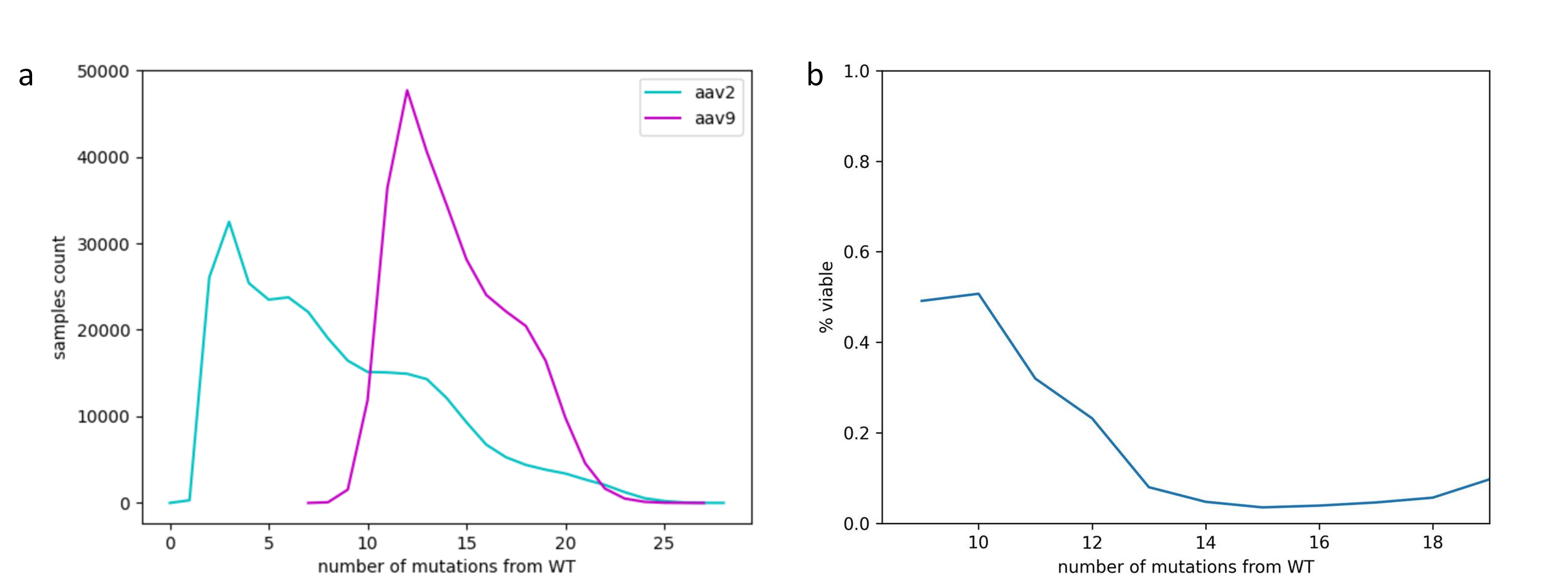}
\caption{\textbf{a:} Proportion of samples generated by the diffusion model at different numbers of mutation sites, where blue represents AAV2 and red represents AAV9. \textbf{b:} Proportion of viable samples for sequences generated by the diffusion model at different numbers of mutation sites. The x-axis represents different numbers of mutation sites, and the y-axis represents the proportion of viable samples.}
\label{fig:results_pdAAV9}
\end{figure}

Previously, the only available approach for sequence design based on a specific serotype of AAV capsid, where no known viable mutant sequences exist, was random mutation design. However, previous findings on AAV2 \cite{ogden2019comprehensive} have revealed that the proportion of viable samples decreases significantly, reaching nearly zero, when the number of mutations exceeds five. Due to the high similarity in capsid sequence between AAV9 and AAV2, we aimed to test the effectiveness of transferring sequences generated by a model trained on AAV2 data to the wild-type AAV9 at corresponding positions. The experimental results shown in \cref{fig:results_pdAAV9}a indicate a significant increase in the number of mutations when the sequences generated from the corresponding region of AAV2 were transferred to the corresponding region of AAV9, as compared to the wild-type AAV9 sequence. \cref{fig:results_pdAAV9}b demonstrates that the proportion of generated sequences that remained viable in AAV9 was approximately 50\% for mutation numbers ranging from 9 to 10. Conversely, when employing random mutation methods (as referenced from dyno data) for sequence mutation in AAV9 without any viability labeling data, the proportion of viable samples was close to zero at a mutation number of 9. Additional detailed information regarding the viability proportions can be found in \cref{tab:dataset_AAV9}. These findings indicate the potential utilization of existing data from other serotypes to create diverse models for sequence design in future serotype capsid design, instead of solely relying on random mutation approaches.

\subsection*{Analyzing Mutations in AAV9 Hypervariable Regions.}\label{subsec:results_prokaryotic}

Apart from investigating hypervariable region VIII (HVR VIII), our study focused on exploring the extensively studied regions of HVR IV and HVR V within the AAV9 capsid. These regions have been recognized for their ability to tolerate mutations, and our objective was to gain a comprehensive understanding of their mutational landscape.

\begin{figure}[h!]
\centering
\includegraphics[width=0.99 \textwidth]{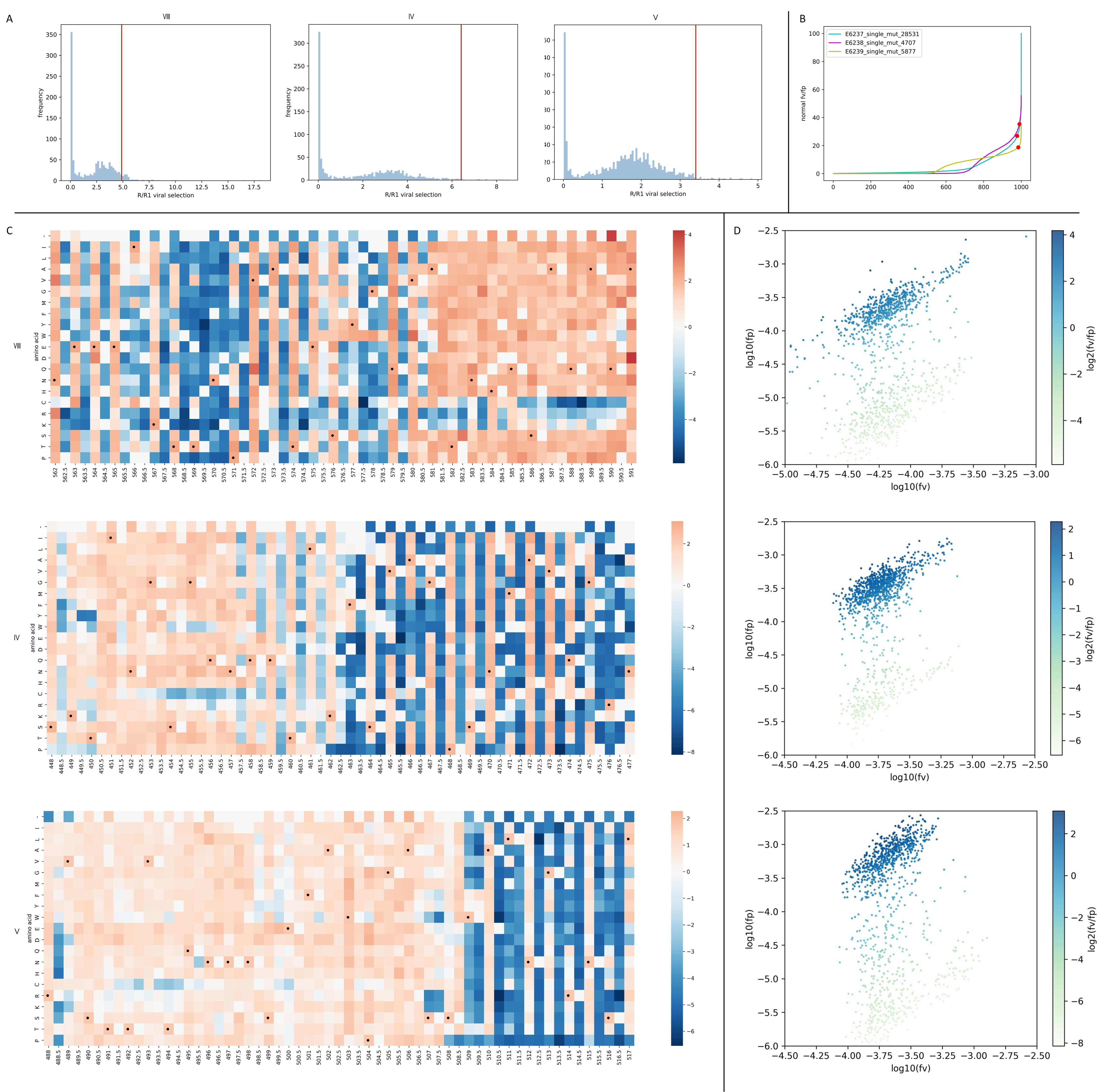}
\caption{\textbf{a:} Distribution of activity values for single-mutant sequences. The x-axis represents the activity values of the sequences, and the y-axis represents the frequency of sequences within that activity value range. \textbf{b:} Trend of activity values for single-mutant sequences. The x-axis represents each sequence, and the y-axis represents the normalized activity values. \textbf{c:} Enrichment levels of single-mutant sequences at different mutation positions. From top to bottom are the enrichment levels of sequences with single mutations in regions VIII, IV, and V of AAV9. The x-axis represents the mutation positions in the current region. If the mutation position is a float value, it indicates an insertion between two integer positions. The y-axis represents the amino acid type after the mutation, where "-" represents the deletion of the amino acid at the current position. Black dots represent the positions of wild-type sequences in that region. \textbf{d:} Enrichment levels of single-mutant sequences. From top to bottom are the enrichment levels of sequences with single mutations in regions VIII, IV, and V of AAV9. The x-axis represents the frequency of plasmids after sequencing, and the y-axis represents the frequency of viruses after sequencing.}
\label{fig:results_bio}
\end{figure}

In particular, our focus was on amino acid residues 448-476, 488-517, and 562-590, where we performed single amino acid mutations within these regions. Furthermore, we introduced random amino acid insertions between adjacent residues. To evaluate the viability of these mutations, we calculated activity values by comparing the frequency of the vector to the frequency of the plasmid, as depicted in \cref{fig:results_bio}a. The red line on the graph signifies the activity value of the wild-type sequence.

The analysis indicated that the majority of peak reads were concentrated between 0 and 0.5, implying that sequences within this range were either inviable or demonstrated reduced viability. The enriched capsid sequences exhibited a distribution that resembled a Gaussian curve.

Upon comparing the activity levels of HVR IV and HVR V with those of HVR VIII, it became apparent that HVR VIII demonstrated both a higher activity peak value and a broader range (1-6 compared to 1-3). The frequency comparisons in \cref{fig:results_bio}b revealed that while HVR V variants exhibited reads in 80\% of cases, HVR IV and HVR VIII variants had reads in only 60\% of cases. However, HVR IV and HVR VIII variants exhibited a greater number of variants with significantly higher reads compared to HVR V mutants, with HVR VIII mutants demonstrating the highest read counts. The wild-type sequences were indicated as red dots. Consequently, HVR V encompassed a broader range of variability, while HVR IV and HVR VIII variants exhibited the highest read counts.

In \cref{fig:results_bio}c, we presented a comprehensive breakdown of fitness scores for the insertion, deletion, and mutation of each amino acid within the selected HVR regions. These scores were calculated based on the logarithm base 2 of the vector frequency divided by the plasmid frequency. The score ranges for HVR VIII, HVR IV, and HVR V mutants were approximately [-5 to 4], [-8 to 2.5], and [-6.5 to 2], respectively. These scores substantiated that HVR VIII comprised a subset of variants with superior fitness scores. Importantly, we identified the regions of mutation and insertion tolerance, specifically spanning amino acid residues 588-591, 448-462, and 488-508.

Intriguingly, HVR V demonstrated the most extensive tolerant region, indicating the need for further investigation into more substantial mutations within this region. \cref{fig:results_bio}d illustrated the vector and plasmid frequency at each variant level, unveiling a distinct clustering of variant populations into two clusters, with minimal neutral mutations.

\section*{Discussion}

In this study, we employed the diffusion model to generate sequences within region VIII of AAV2 and conducted activity experiments on AAV2 capsids. The results revealed that the generated sequences displayed a viability proportion exceeding 90\% within the range of 7 to 20 mutations. This finding highlights the robust capability of our model in generating viable sequences.

Additionally, we utilized the diffusion model to generate sequences within region VIII of AAV2 and performed activity experiments on AAV9 capsids. The results unveiled that the generated sequences displayed a viability proportion of approximately 50\% when the number of mutations ranged from 9 to 10. This proportion was notably higher than the viability proportion obtained through random mutation-based sequence design in the absence of viable sequences.

Traditionally, the experimental process for designing capsids with a higher number of mutation sites involved initial experiments utilizing single-site mutagenesis, followed by rational and random mutagenesis based on the obtained results. This iterative process aimed to generate additional experimental data, ultimately leading to the discovery of a broader range of capsid sequences. Based on the results presented in this paper, our model can be utilized for the design of capsids for different AAV serotypes, obviating the need for random mutation or exhaustive single-site mutagenesis. This expedites the experimental process for AAV capsid design.

However, our model has certain limitations. One notable limitation is that the range of mutation counts for the generated sequences is constrained by the range observed in the training set. o overcome this limitation, future improvements can involve pre-training the model on an expanded dataset comprising not only AAV capsid sequences but also sequences from other viruses and even non-viral protein sequences. By enabling the model to generate high-quality protein sequences, we can subsequently perform astute fine-tuning on the existing viable AAV samples, liberating the model from the constraints of the AAV training set and unleashing the capabilities acquired through pre-training.

In \cref{fig:results_bio}c, the heat map suggests that the mutant-tolerant region may extend beyond the range of our tests. Expanding the scope of saturated mutagenesis could provide further valuable insights. Notably, distinct patterns emerged within each HVR domain. In the aforementioned tolerance regions, amino acids K, R, and C were found to be unfavorable in HVR VIII, likely due to their large size and potential disruption of the capsid structure. In HVR IV (residues 457-475), direct mutations were better tolerated than insertions, highlighting the importance of residue length and structural rigidity in this region. While we gathered data on multiple mutations for HVR VIII, obtaining similar data for HVR IV and V regions would be beneficial. Additionally, collecting data on double or multiple mutation/insertion/deletion scenarios could unveil synergistic effects on fitness and introduce new factors that impact vector transduction.

This study marks a significant advancement in capsid engineering, highlighting the correlation between VP sequence mutants and capsid assembly features through an innovative algorithm. By enabling the investigation of transduction efficacy and specificity predictions, our research offers valuable insights into the field of capsid design, where the transduction function is intricately linked to the structure of the vector capsid, which is determined by the VP sequence. Consequently, it is crucial to establish a robust selection model for data collection and algorithm development to further explore these captivating areas of study.

\section*{Methods}
\label{sec:methods}

\subsection*{The process of sequence generation using the diffusion model}\label{subsec:paralog_matching_problem}

The task entails generating sequences within the mutation region of the AAV2 capsid, specifically targeting region VIII. Initially, we compiled mutation sequences from dyno in this region, forming the training set for our model \cite{bryant2021deep}. The dataset comprised a total of 140,000 data entries, encompassing a range of mutation site numbers from 1 to 28. The capsid sequence consists of multiple amino acids, with each amino acid regarded as a token. Hence, we utilize a discrete diffusion generation model for sequence generation.

\begin{figure}[h!]
\centering
\includegraphics[width=0.6 \textwidth]{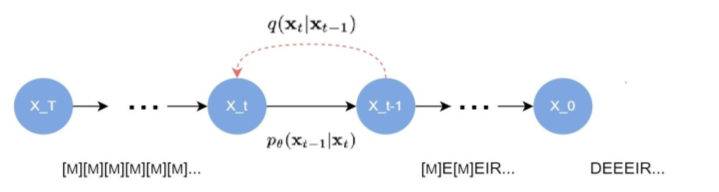}
\caption{\textbf{The directed graphical model considered in this work}}
\label{fig:methods_diff}
\end{figure}

As depicted in \cref{fig:methods_diff}, it presents the implementation diagram of the generation diffusion model \cite{ho2020denoising}. The model consists of two processes: diffusion and denoising. The denoising process can be perceived as a prediction process. Utilizing a maximum length noise sequence, the denoising model trained during the diffusion process progressively eliminates noise from the input sequence. After T steps, the noise sequence is restored to a valid sequence. In the case of a valid sequence, noise is incrementally introduced step by step, producing sequences $x_1,x_2,\ldots$, ultimately resulting in a fully noisy sequence $x_T$. The purpose of the diffusion process is to aid the neural network in acquiring knowledge of the denoising process. Throughout this process, we possess the actual sequences along with the outcomes of adding noise to them. Consequently, this enables the network to learn a mapping that can restore the original sequence from the noisy counterpart. The complete implementation process is subdivided into four stages: data augmentation, noise addition, model training, and denoising. For more detailed information, please refer to the supplementary materials. Once the model is trained, when presented with a fixed-length noise sequence, it progressively recovers the noise sequence to a meaningful and valid capsid sequence.

\subsection*{Generation of AAV Capsid Libraries}\label{subsec:aav_library}

In the development of AAV capsid libraries, wild-type cap genes were subject to modification through the incorporation of DNA oligonucleotides, the sequences of which are provided in Supplementary Table [Insert Table Number]. The specific synthesis of 84-mer to 108-mer DNA oligonucleotides, encoding peptides of interest, was performed by Twist Bioscience in a chip-based primer pool. Subsequently, these DNA oligonucleotides underwent amplification through PCR, utilizing a high-fidelity DNA polymerase (NEB). The resulting PCR fragments were then ligated into the AAV backbone plasmid. The ligation products were transformed into electrocompetent cells (Lucigen) to enhance transformation efficiency. The capsid library plasmids were ultimately prepared using a QIAGEN kit, and the diversity of the capsid library was characterized through next-generation sequencing analysis.

\subsection*{AAV Production Assay and Virus Titer Detection}\label{subsec:aav_product }

To produce viral particles, the plasmid libraries were transfected into 293TN cells. These cells were maintained in a sterile environment within a 5\% CO2 incubator at 37$^{\circ}$C. Typically, 293TN cells were cultured in High-Glucose Dulbecco’s Modified Eagle's Medium (DMEM; Gibco) supplemented with 10\% fetal bovine serum (FBS; Gibco) and 1\% penicillin/streptomycin (Thermo Fisher). AAV library vectors were produced by transfecting 293TN cells, along with adenovirus helper and AAV Rep-$\triangle$Cap plasmids, using FectoVIR (Polyplus). For the transfection process, 293TN cells were seeded in 10cm dishes at a density of 7.2×106 cells per dish. Following a 72-hour duration, the virus was harvested and subsequently purified utilizing an iodixanol density gradient ultracentrifugation method. The AAV titers were quantified using Taqman-based qPCR.

\subsection*{Next-Generation Sequencing}\label{subsec: sequencing}

The remaining cap gene sequences in the purified pool represent viable mutants suitable for both capsid assembly and genome packaging. To assess this, the purified capsids were subjected to heat denaturation at 98 °C for 10 minutes. Subsequently, the mutant region of the cap gene was amplified using High Fidelity 2x master mix (NEB) with PCR primer sequences. Illumina sequencing adapters and indices were integrated in a subsequent PCR step. These PCR amplicons were subsequently subjected to sequencing with overlapping paired-end reads employing Illumina NextSeq.

\begin{refcontext}[sorting = none]
\printbibliography
\end{refcontext}


\newpage
\appendix

\begin{center}
\LARGE{\textbf{Supplementary material}}
\end{center}
\vspace{0.2cm}

\renewcommand{\thesection}{S\arabic{section}}
\renewcommand{\thefigure}{S\arabic{figure}}
\setcounter{figure}{0}
\renewcommand{\thetable}{S\arabic{table}}
\setcounter{table}{0}
\renewcommand{\theequation}{S\arabic{equation}}
\setcounter{equation}{0}

\section{Supplementary methods}

\subsection{Sequencing Data Processing and Analysis of Viability Threshold for Sequences}
\label{supp_meth:exp_process}

\begin{figure}[h!]
\centering
\includegraphics[width=0.9 \textwidth]{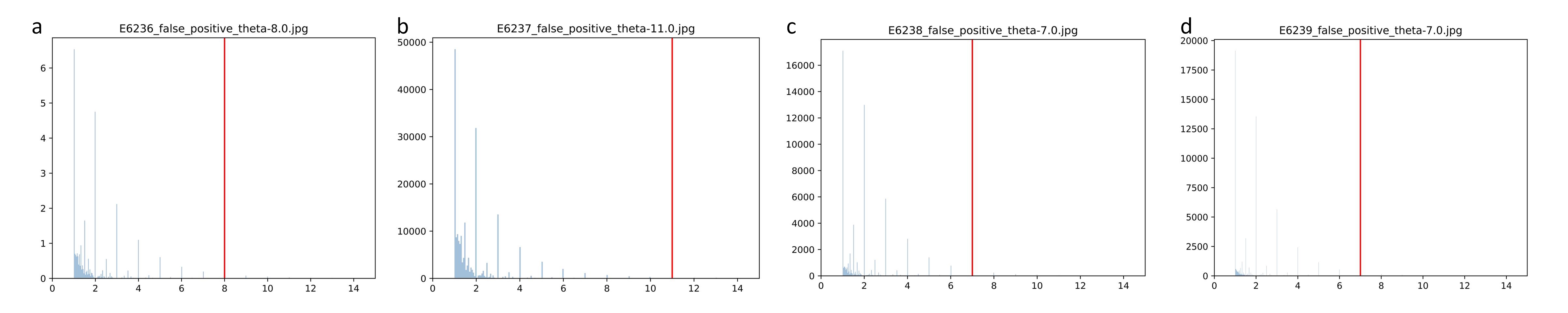}
\caption{\textbf{a-d:} Distribution of the ratio of forward and reverse reads for sequence reads in AAV2 region VIII library, AAV9 region VIII library, AAV9 region IV library, and AAV9 region V library.}
\label{fig:sup_filter1}
\end{figure}

\begin{figure}[h!]
\centering
\includegraphics[width=0.9 \textwidth]{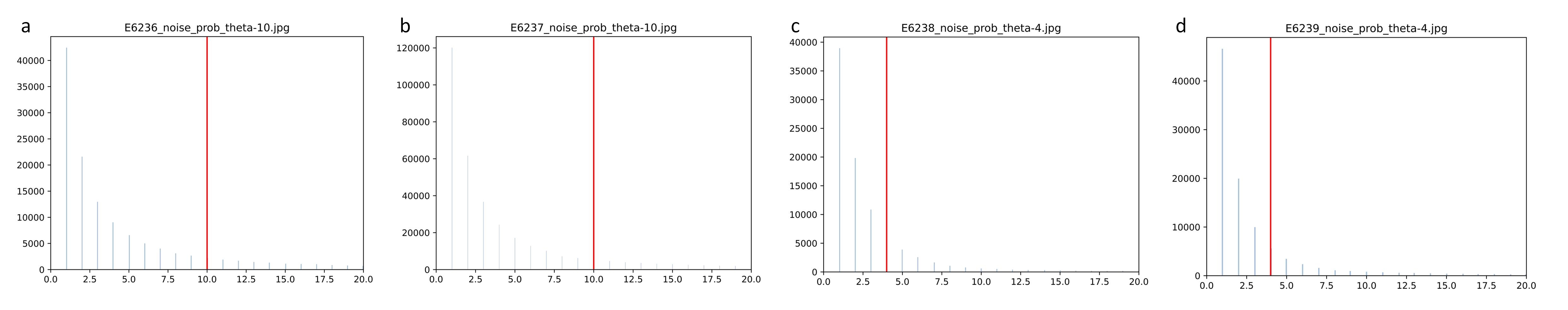}
\caption{\textbf{a-d:} Distribution of viral read counts for AAV2 region VIII library, AAV9 region VIII library, AAV9 region IV library, and AAV9 region V library.}
\label{fig:sup_filter2}
\end{figure}

\begin{figure}[h!]
\centering
\includegraphics[width=0.9 \textwidth]{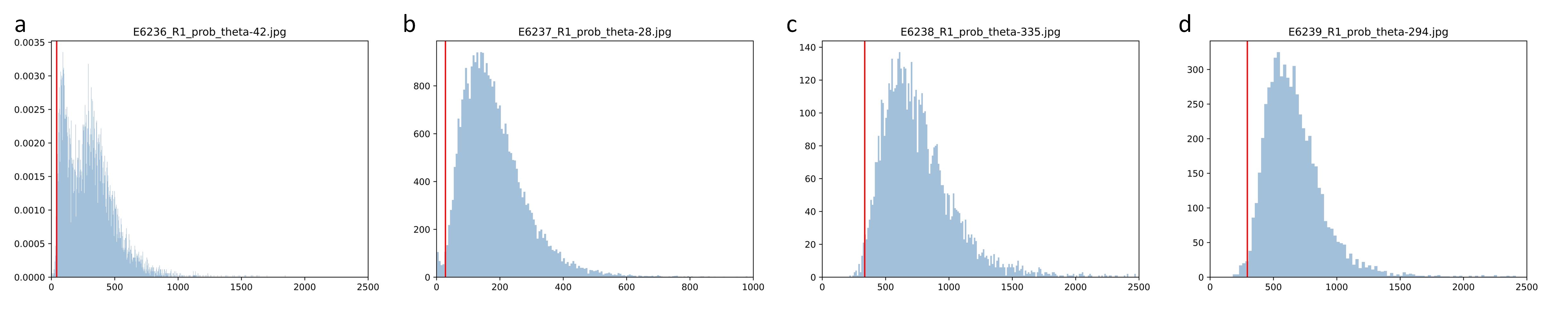}
\caption{\textbf{a-d:} Distribution of plasmid read counts for AAV2 region VIII library, AAV9 region VIII library, AAV9 region IV library, and AAV9 region V library.}
\label{fig:sup_filter3}
\end{figure}

The number of read types obtained from the sequencing results exceeded the range originally designed for the library. To eliminate false positive reads introduced during the experimental/sequencing process, and in addition to filtering out low-quality reads using the default parameters of fastp, we implemented the following supplementary strategies: 1. In the case of plasmid libraries or viral libraries, reads with an extreme imbalance in terms of the forward and reverse read count ratio were filtered out, as illustrated in \cref{fig:sup_filter1}; 2. For viral libraries, reads that fell outside the range designed for the library were filtered out by applying the 80th percentile read count threshold, thereby eliminating all reads below this threshold, as depicted in \cref{fig:sup_filter2}; 3. Only the sequences specifically designed for the library were chosen for subsequent analysis; 4. Abnormal plasmid data was eliminated. Following plasmid packaging, the plasmid sequencing results of certain sequences may exhibit abnormalities or have a scarcity of plasmids to generate viruses. Consequently, we discarded sequences with an insufficient number of plasmids, retaining 99\% of the sequences based on the plasmid distribution, as illustrated in \cref{fig:sup_filter3}.

\begin{figure}[h!]
\centering
\includegraphics[width=0.7 \textwidth]{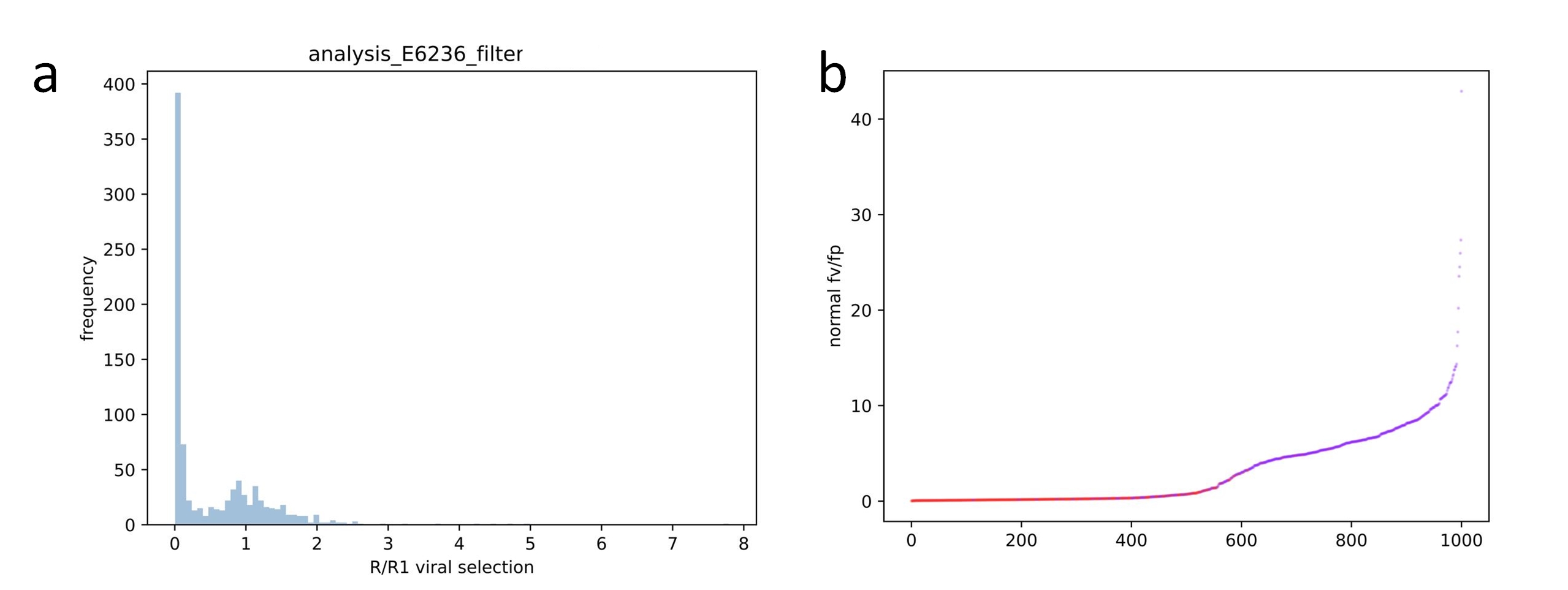}
\caption{\textbf{a:} Distribution of Activity Values. where the x-axis represents the activity values and the y-axis represents the frequency of sequences within that activity value range. \textbf{b:} Trend of Activity. where the x-axis represents the selected 1000 samples and the y-axis represents the normalized activity values. In this figure, red color represents sequences labeled as non-viable in the dyno published dataset, while purple color represents sequences labeled as viable.}
\label{fig:sup_viable_thresh}
\end{figure}

After obtaining clean plasmid and viral sequencing data, the enrichment level of each sequence was calculated as follows: the Python script was utilized to count the number of each variant in the sequencing library (cm). Prior to calculating the frequency, the counts across replicates were summed. The frequency of each variant in the viral library (fv) or plasmid library (fp) was calculated as $f = \frac{cm}{\sum{cm}}$. The enrichment level of each variant in the viral pool was calculated as $s = \frac{fv}{fp}$. After obtaining the enrichment level distribution for each library, various methods were employed to determine the activity threshold for each library based on the characteristics of the respective data. For Experiment 1.1, a control group consisting of 1000 randomly sampled publicly available dyno data was collected to be used in conjunction with Experiment 1.1. The ratio of positive to negative activity samples in the control group was approximately 1:1. Finally, a mixture Gaussian simulation was conducted using these 1000 sets of samples to establish the threshold for the final positive and negative samples, as depicted in \cref{fig:sup_viable_thresh}. For Experiment 1.2 and Experiment 2, saturated single-mutation data was extracted and subjected to a mixture Gaussian simulation to determine the threshold for the final positive and negative samples, as illustrated in \cref{fig:results_bio}a.

\subsection{Diffusion Model for Sequence Generation}
\label{supp_meth:model_process}

\paragraph{Data Augmentation} The lengths of viable mutated sequences vary. To ensure consistent lengths for each sequence and enhance data diversity, we introduce the [del] token to denote the deletion of an amino acid at a specific position. Accordingly, [del] tokens are randomly inserted at positions within the mutated sequence, ensuring that the final sequence length aligns with the specified maximum sequence length.

\paragraph{Noise Injection} The diffusion model is categorized into continuous diffusion and discrete diffusion \cite{sohl2015deep}, with the latter being more appropriate for capsid amino acid mutation design. Hence, this paper leverages the algorithm of the D3PM diffusion model \cite{austin2021structured}. Pre-defined transition probability matrices, denoted as $Q_t$, are employed to ascertain the transition probabilities between amino acid types at various time steps. Specifically, $[Q_t]{mn}=q(x_t=m|x_{t-1}=n)$ represents the probability of transitioning from amino acid type n at time step t-1 to amino acid type m at time step t. The matrix $Q_t$ can be defined arbitrarily, as long as the column sums add up to 1. Using $Q_t$ and the initial state $x_0$, we can generate noisy data $x_t$ at any given time step. The calculation is performed using the formula $q(x_t|x_0)=v^T(x_t)\bar{Q}_tv(x_0)$, where $\bar{Q}_t=Q_t\cdot\cdot\cdot Q_1$ and $v(x)$ represents the one-hot vector with a value of 1 at state x and 0 for all other states.

\begin{equation}
\boldsymbol{Q_t}=\begin{bmatrix}\alpha_t+\beta_t&\beta_t&\beta_t&\cdots&0\\\beta_t&\alpha_t+\beta_t&\beta_t&\cdots&0\\\beta_t&\beta_t&\alpha_t+\beta_t&\cdots&0\\\vdots&\vdots&\vdots&\ddots&\vdots\\\gamma_t&\gamma_t&\gamma_t&\cdots&1\end{bmatrix}
\end{equation}

\paragraph{Model Training} The model structure employed in this study is the encoder component of the Transformer \cite{vaswani2017attention}. The input to the model is the noisy sequence at the current time step, while the output represents the denoised viable initial sequence, denoted as $p_\theta(x_0|x_t)$. The training of this model involves the utilization of the following loss function:

\begin{equation}
\begin{aligned}L_\mathrm{vb}=\mathbb{E}_{q(\boldsymbol{x}_0)}\bigg[\underbrace{D_\mathrm{KL}[q(\boldsymbol{x}_T|\boldsymbol{x}_0)||p(\boldsymbol{x}_T)]}_{L,T}+\sum_{t=2}^T\underbrace{\mathbb{E}_{q(\boldsymbol{x}_t|\boldsymbol{x}_0)}\big[D_\mathrm{KL}[q(\boldsymbol{x}_{t-1}|\boldsymbol{x}_t,\boldsymbol{x}_0)||p_\theta(\boldsymbol{x}_{t-1}|\boldsymbol{x}_t)\big]\big]}_{L_{t-1}}\\\underbrace{-\mathbb{E}_{q(\boldsymbol{x}_1|\boldsymbol{x}_0)}\big[\log p_{\theta}(x_0|x_1)]}_{L_0}\bigg]\end{aligned}
\end{equation}

\begin{equation}
\mathrm{p_\theta(x_{t-1}|x_t)=\sum_{x_0}p(x_{t-1}|x_t,x_0)p_\theta(x_0|x_t)}
\end{equation}

\begin{equation}
\begin{aligned}q(x_{t-1}|x_t,x_0)&=\frac{q(x_t|x_{t-1},x_0)q(x_{t-1}|x_0)}{q(x_t|x_0)}\\&=\frac{\left(\boldsymbol{v}^\top(x_t)\boldsymbol{Q}_t\boldsymbol{v}(x_{t-1})\right)\left(\boldsymbol{v}^\top(x_{t-1})\overline{\boldsymbol{Q}}_{t-1}\boldsymbol{v}(x_0)\right)}{\boldsymbol{v}^\top(x_t)\overline{\boldsymbol{Q}}_t\boldsymbol{v}(x_0)}.\end{aligned}
\end{equation}

As the number of time steps $T$ approaches infinity, both the forward process and the reverse process exhibit the same functional form \cite{feller2015retracted}, Because $q(x_T|x_0)$ converges to a stationary distribution as t approaches infinity, the term $L_T$ can be disregarded. Consequently, the minimization of $\sum^T_{t=2}L_{t-1}+L_0$ leads to model training.

\paragraph{Denoising} The denoising process entails obtaining data with reduced noise at the previous time step, relying on the data with noise at the current time step. This relationship is as $p_\theta(x_{t-1}|x_t)$. The computation of this value is as follows:

\begin{equation}
\mathrm{p_\theta(x_{t-1}|x_t)=\sum_{x_0}q(x_{t-1}|x_t,x_0)p_\theta(x_0|x_t)}
\end{equation}

\begin{figure}[h!]
\centering
\includegraphics[width=0.9 \textwidth]{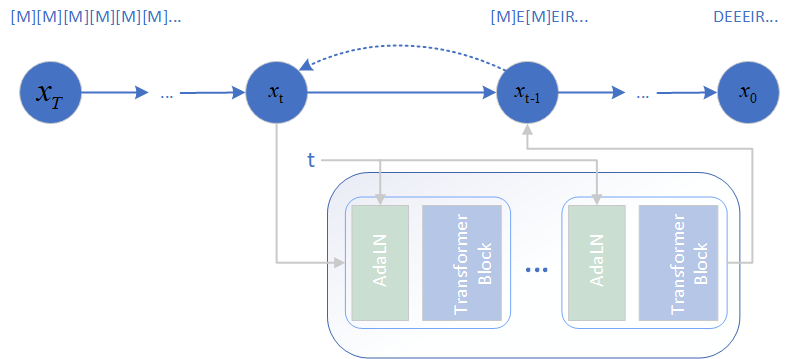}
\caption{Overall framework of our method}
\label{fig:sup_framework}
\end{figure}

\paragraph{Model Architecture} The model framework is shown in \cref{fig:sup_framework}, and the core model consists of 12 layers of transformer blocks. The parameters for each block in each layer are as follows:

\begin{itemize}[noitemsep,topsep=0pt]
    \item Input shape: (56, 21)
    \item hidden size: 512
    \item num attention heads: 16
    \item intermediate size: 4096
    \item hidden act: gelu
\end{itemize}

Other training parameters are as follows:

\begin{itemize}[noitemsep,topsep=0pt]
    \item warmup ratio: 0.1
    \item learning rate: 1e-4
    \item weight decay: 0.04
    \item lr scheduler type: linear
\end{itemize}

\section{Supplementary tables}

\begin{table}[htbp]
    \centering
    \begin{tabular}{c|c|c}
    \toprule
    \textbf{sequence count}&\textbf{ \% viable}&\textbf{num mutations}
    \\ \hline
    1772&0.511561175&4\\
    495&0.7625711122&5\\
    500&0.93144&6\\
    500&0.932&7\\
    509&0.931784&8\\
    499&0.94591784&8\\
    499&0.57785812&9\\
    499&0.969558012&10\\
    470&0.962561702&11\\
    482&0.931227861&12\\
    499&0.96169429&15\\
    499&0.99629429&15\\
    297&0.99525695&15\\
    267&0.996662718&16\\
    267&0.98632718&16\\
    298&0.9385323&17\\
    298&0.93576287&16\\
    178&0.9567457&16\\
    17&0.5662969&20\\
    69&0.56229609&20\\
    7&0.952742927&22\\
    1&1&23 \\
    \hline
    \bottomrule
    \end{tabular}
\caption{Performance on AAV2}
\label{tab:dataset_AAV2}
\end{table}

\begin{table}[htbp]
    \centering
    \begin{tabular}{c|c|c}
    \toprule
    \textbf{sequence count}&\textbf{ \% viable}&\textbf{num mutations} 
   \\ \hline
    1321& 0.491294474 & 9 \\
    1696& 0.507075472 & 10  \\
    1583& 0.319646241 & 11  \\
    6527& 0.231806343 & 12  \\
    1356& 0.079646018 & 13  \\
    1394& 0.047345768 & 14  \\
    1398& 0.035050072 & 15  \\
    1488& 0.038978495 & 16  \\
    1568& 0.045918367 & 17  \\
    1522& 0.056504599 & 18  \\
    1598& 0.096996245 & 19  \\
    1299& 0.344880677 & 20  \\
    417& 0.323741007 & 21 \\
    89& 0.393258427 & 22 \\
    11&0.181818182&23  \\
    \hline
    \bottomrule
    \end{tabular}
\caption{Performance on AAV9}
\label{tab:dataset_AAV9}
\end{table}

\end{document}